\documentclass{article}
\usepackage{PRIMEarxiv}
\usepackage{amsfonts}       
\usepackage{nicefrac}       
\usepackage{microtype}      
\usepackage{fancyhdr}       
\usepackage{hyperref}

\usepackage{parcolumns}
\usepackage{booktabs}
\usepackage{multirow}

\usepackage{listings,newtxtt}
\usepackage{xcolor}
\usepackage{graphicx}
\usepackage{physics}
\usepackage{bm}
\usepackage{amsmath, amssymb}
\usepackage{geometry}
\newsavebox{\dotbox}
\newcommand{\outerdot}[1]{%
\sbox\dotbox{$#1$}\dot{\usebox\dotbox}}
\usepackage{cleveref}
\usepackage{natbib}
\definecolor{codegreen}{rgb}{0,0.6,0}
\definecolor{codegray}{rgb}{0.5,0.5,0.5}
\definecolor{codepurple}{rgb}{0.58,0,0.82}
\definecolor{backcolour}{rgb}{0.95,0.95,0.92}

\lstdefinestyle{mystyle}{
    backgroundcolor=\color{backcolour},
    commentstyle=\color{codegreen},
    keywordstyle=\color{magenta},
    numberstyle=\tiny\color{codegray},
    stringstyle=\color{codepurple},
    basicstyle=\ttfamily\footnotesize,
    breakatwhitespace=false,
    breaklines=true,
    captionpos=b,
    keepspaces=true,
    numbers=left,
    numbersep=5pt,
    showspaces=false,
    showstringspaces=false,
    showtabs=false,
    tabsize=2
}
\newcommand{\swarmrl}{\texttt{SwarmRL}}

\DeclareMathOperator*{\argmax}{arg\,max}

\title{SwarmRL: Building the Future of Smart Active Systems}

\author{
  Samuel Tovey\footnotemark[2], Christoph Lohrmann\footnotemark[2], Tobias Merkt \And David Zimmer, Konstantin Nikolaou, Simon Koppenh\"{o}fer \And Anna Bushmakina, Jonas Scheunemann,  Christian Holm \\
  Institute for Computational Physics \\
  University of Stuttgart \\
  70569, Stuttgart, Germany\\
  \texttt{\{stovey, clohrmann, holm\}@icp.uni-stuttgart.de} \\
}

\begin{document}
\maketitle
\footnotetext[2]{These authors contributed equally}
\begin{abstract}
This work introduces \swarmrl{}, a Python package designed to study intelligent active particles.
\swarmrl{} provides an easy-to-use interface for developing models to control microscopic colloids using classical control and deep reinforcement learning approaches.
These models may be deployed in simulations or real-world environments under a common framework.
We explain the structure of the software and its key features and demonstrate how it can be used to accelerate research.
With \swarmrl{}, we aim to streamline research into micro-robotic control while bridging the gap between experimental and simulation-driven sciences.
\swarmrl{} is available open-source on GitHub at \url{https://github.com/SwarmRL/SwarmRL}.
\end{abstract}

\section{Introduction}
\label{sec:introduction}
Mastering control of microrobots at the microscopic scale has the potential for insights and the development of new technologies capable of changing the world.
Whether it be an improved understanding of bacterial navigation strategies or direct control over microscopic robots, numerous fields, including construction~\citep{hsu16a}, plant pollination and ecosystem defense~\citep{stefanec22a}, and search and rescue~\citep{murphy08a}, will be enhanced by advancements in micro-scale control research.
One field in particular that will see significant changes is medicine, where it is expected that micro robotic agents will be capable of advanced treatment strategies including targeted cancer therapies~\citep{schmidt20a}, assisted and drug delivery~\citep{nelson23a, felfoul16a, hosseinidoust16a, zhuang15a}, assisted fertilisation~\citep{medina16a, khalil14a}, amongst many others~\citep{nelson10a, li22a}.
The two most significant challenges in realising this future are developing the tools and materials to build such capable microscopic agents and programming them to achieve complex tasks with minimal observed input.
These agents should not only be capable of performing complex actions such as swimming, pushing, and perhaps, communications, but they should do so without constant input from a supervisor.
The first of these challenges has been, and is being, extensively investigated.
Currently, there are many approaches for designing mobile micro robots~\citep{dabbagh22a}.
Initial approaches included the light-driven Janus particles~\citep{su19a} and coils~\citep{wang19a} and magnetically driven devices taking on numerous shapes~\citep{shen23a, dhatt23a}.
Recent approaches have also included using living cells to produce motion on this microscopic scale, the Xenobots being a prime example~\citep{kriegman20a, blackiston21a}.
However, the control of these agents, once built and deployed, is a complicated problem.
Due to the size of these agents, directly installing computational processing power onto them is, thus far, intractable.
Therefore, simplistic algorithms that aim to encode reactionary behaviour into the agents are often constructed, something that would require limited on-board processing.
Examples of this include the use of microrobots to perform chemotaxis~\citep{hartl21a} by changing their shapes upon exposure to varying fields, object manipulation~\citep{kim16a} through the use of an external magnetic field, and reproducing swarming behaviour by applying simple rotations and translations to individual agents by a set of rules triggered by changes in local environment~\citep{baeuerle20a, lavergne19a}.
More recent approaches have utilised reinforcement learning approaches to further push in the direction of learned control strategies.
\citet{qin23a} utilised a Q-learning algorithm to learn swimming strategies in gated microswimmers,~\citet{borra22a} used an actor-critic framework to learn competitive capture-evasion strategies from hydrodynamic cues, and~\citet{landin21a} investigated actor-critic learned navigation strategies under the influence of stochastic environments.
In all cases, demonstrating under what conditions microrobots can perform complex tasks in various environments brings us a step closer to realising the full potential of this technology.

It is clear that research in the direction of micro-robotic control is both promising and excitingly multi-disciplinary; however, due to the nature of the technology required, whether it be complex experimental equipment, deep knowledge of state-of-the-art machine learning algorithms and their implementation, or the expertise to construct physically realistic simulations, the entry barrier can be high for specialists who focus on only one of these areas.
For this reason, we have developed the \swarmrl{} Python package to combine each control workflow element under a common framework and accelerate algorithm development and deployment.
\swarmrl{} enables researchers to apply control strategies to particles in simulation or experiments.
These strategies can be built from sets of rules that are rigidly programmed and utilise the environment of the particles, or they can be trained using the actor-critic reinforcement learning approach, offering complete flexibility in how the agents are controlled, what actions they are capable of, and what tasks they must perform.
Beyond the customizability, wherever possible, it is built on top of the JAX~\cite{bradbury18a} ecosystem to maximise performance and is suited for deployment on large distributed compute clusters.
This paper aims to introduce the \swarmrl{} software and demonstrate how it can be applied to perform frontier research.
We begin with an overview of the theory underpinning the software, explicitly discussing microscale active matter, simulating it in a physically realistic manner, and the reinforcement learning we have built to control it.
Afterwards, the architecture of the software is described, detailing how each piece of the software fits together and what can be done with them.
Finally, we discuss some of the unique features of \swarmrl{}, including performance and visualisation capabilities, before concluding with an outlook of the project.
With \swarmrl{}, we hope to enable scientists from diverse backgrounds to contribute to and develop the field of micro-robotics research and realise the potential of this technology.
\section{Theory}
\label{sec:theory}
\swarmrl{} harnesses tools from a range of fields, the largest of which are active matter simulation and reinforcement learning.
In order to understand and better utilize the software infrastructure, it is important to also have an understanding of the theory it implements. 
In this section, we introduce the important theoretical concepts on which \swarmrl{} is built, before exploring the architecture of the package.
\subsection{Active Matter}
\label{subsec:active-matter}
\subsubsection*{Biological and artificial active systems}
Active matter refers to systems that consume energy from their surroundings and are thus internally driven out of equilibrium~\citep{romanczuk12a}.
All biological and human-made machines fall under this term. 
However, in the context of this work, we focus on the branch of active systems that consists of small active particles that use energy to perform persistent, directed motion.
This much more narrow definition covers plenty of examples in biology and engineering.
Many bacteria like Escherichia coli~\citep{wadhwa22a}, archaea like Thermoplasma volcanium~\citep{jarrell21a}  and small eukaryotes like Chlamydomonas reinhardtii~\citep{silflow01a} use molecular motors and organelles like archaella, flagella or pili to self-propel.
Artificial microswimmers can be made from colloidal particles that use, e.g., catalytic reactions~\citep{howse07a}, self-generated temperature gradients~\citep{jiang10a} or magnetic fields~\citep{mandal18a} to generate propulsion.
In some more rare cases, the active particles can self-propel and self-steer, i.e., use torque to actively change their direction.
For bacteria, tumbling~\citep{berg72a, darnton07a} is a well-known mechanism for change of direction.
Some artificial microswimmers can also be steered by, e.g., change in laser focus~\citep{baeuerle20a} or magnetic field direction~\citep{carlsen14a}.
In all these examples, the active particle is of micrometer size and suspended in a liquid, making translational and rotational Brownian motion a non-negligible factor that competes with the deterministic active motion.

\subsubsection*{Simulation}
Active matter systems with self-propelled small particles are usually modelled within the framework of stochastic dynamics of the individual constituents. 
We use the two- or three dimensional overdamped Langevin equations
\newcommand{\partclPos}{\vb{r}}
\newcommand{\partclDirec}{\vb{\hat{e}}}
\newcommand{\partclNormal}{\vb{\hat{n}}}
\newcommand{\partclAngVel}{{\boldsymbol{\omega}}}
\newcommand{\boltzmannKonstant}{k_\text{B}}
\newcommand{\noise}{\boldsymbol{\xi}}
\newcommand{\activeForce}{F^\text{act}}
\newcommand{\activeTorque}{M^\text{act}}

\begin{equation}
    \dot{\partclPos}_i(t) = \gamma_t^{-1} \left[\activeForce_i(t) \partclDirec_i(t) + \vb{F}_i(\partclPos_i, \qty{\partclPos_j}) \right] + \sqrt{2 \boltzmannKonstant T \gamma_t^{-1}} \noise^t_i(t),
    \label{eq:langevin_pos}
\end{equation}
\begin{equation}
    \outerdot{\partclDirec}_i(t) = \left[\gamma_r^{-1} \left[\activeTorque_i(t) \partclNormal_i(t) + \vb{M}(\partclPos_i, \qty{\partclPos_j}) \right] + \sqrt{2 \boltzmannKonstant T \gamma_r^{-1}} \noise^r_i(t)\right] \cross \partclDirec_i(t),
    \label{eq:langevin_direc}
\end{equation}
where $\partclPos_i$ denotes the position of particle $i$, $\gamma_{t(r)}$ the translational (rotational) friction coefficient, $\activeForce$ the active driving force that models self-propulsion, $\partclDirec$ the unit vector representing the direction of the particle, $\vb{F}_i(\partclPos_i, \qty{\partclPos_j})$ a conservative force from interactions between particle $i$ and its environment as well as from interactions with other particles $j$, $\boltzmannKonstant$ the Boltzmann constant, $T$ the absolute temperature, $\noise^{t,(r)}$ the translational (rotational) noise with $\expval{\noise^{t,(r)}_i(t)} = 0 $ and $\expval{\noise^{t,(r)}_i(t)\otimes\noise^{t,(r)}_j(t')} = \delta_{ij}\delta(t-t')\mathbf{1}$, where $\expval{\cdot}$ denotes an ensemble average (expectation value) and $\mathbf{1}$ the unity matrix, $\activeTorque$ an active steering torque, $\partclNormal$ a unit vector perpendicular to $\partclDirec$ that sets the steering direction and $\vb{M}_i(\partclPos_i, \qty{\partclPos_j})$ an interaction torque analogous to $\vb{F}_i$.
We stress the time dependence of $\activeForce$ and $\activeTorque$ because \swarmrl{} is designed to focus on control and decision-making on the level of single particles.
Collective behavior and task fulfillment are to be achieved by the particles' actions and what they can control rather than by external fields acting on all particles.

\subsection{Reinforcement Learning}\label{subsec:reinforcement-learning}
As a sub-field of ML, RL is concerned with how intelligent agents can take actions in an environment to maximize a cumulative reward. 
It allows these agents to learn optimal protocols to achieve a given task or a set of tasks without explicitly telling them how. 
This is oftentimes useful when a task is too complex for a simple algorithm or control system.
It can also be used when the focus of the investigation is the emergence of a strategy for a problem.
RL is, however, data-inefficient and is thus used mostly for problems where data exists abundantly and where the action space is relatively small such as autonomous driving~\citep{kiran21a}, video games~\citep{mnih15a}, and robotics~\citep{ibarz21a}. 
It reached great popularity in 2016 when Deepmind's AlphaGo won against the top Go player Lee Sedol~\citep{silver16a}.
When using it for specific tasks, however, RL can achieve remarkable results. 
In 2022, it solved a 50-year-old open question in mathematics how to most efficiently use matrix multiplication~\citep{fawzi22a}; this discovery likely improves the efficiency of machine learning significantly due to its widespread use of matrix multiplication. 
It is further increasingly used in Physics research for sophisticated tasks such as quantum error correction~\citep{zeng23a} and the magnetic control of tokamak plasmas~\citep{deepmind_plasma}. 
This section provides an introduction to RL and explains the training approach currently implemented in SwarmRL, namely, deep Actor-Critic (AC).

\subsubsection{Actor-Critic Reinforcement Learning}
\begin{figure}
    \centering
    \includegraphics[width=\linewidth]{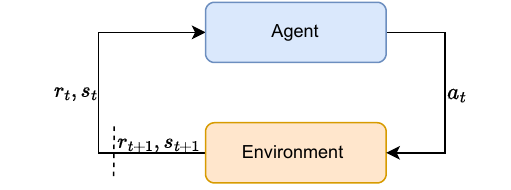}
    \caption{Graphical overview of the RL workflow. The agent selects an action, $a_{t}$, based in its current state, $s_{t}$, and acts within the environment. This action yields a new state, $s_{t+1}$ and an associated reward, $r_{t+1}$.}
    \label{fig:rl-workflow}
\end{figure}

The RL training paradigm is concerned with iteratively improving an agent's decision-making such that it can achieve a pre-defined task as efficiently as possible.
In practice, this is performed by placing agents into an environment with which they can interact by performing actions, $a_{t}$.
To decide on such an action, the agent receives a state description, $s_{t}$, that describes the environment with sufficient information.
The model used to decide on the action is referred to as the policy, $\pi_{\theta}: s_{t} \rightarrow a_{t}$, often taken to be a neural network with parameters, $\theta$ that is then referred to as an \texttt{actor}.
These actors typically produce a probability distribution, which can then be sampled to select an action given the state.
After an action is taken, a new state is produced along with a reward, quantifying the agent's progress towards achieving its task, a process outlined in Figure~\ref{fig:rl-workflow}.
Fundamentally, the goal of training in RL is to find the optimal policy that, throughout a trajectory, $\tau$, maximises a cost function, $J(\pi_{\theta})$, i.e.,
\begin{equation}
    \pi^{*} = \argmax_{\pi} J(\pi_{\theta}) = \argmax_{\pi} \int_{\tau}P\left(\tau | \pi_{\theta}\right)\cdot R(\tau) = \argmax_{\pi} \left\langle R(\tau) \right\rangle_{\tau \sim \pi_{\theta}},
\end{equation}
where $R(\tau)$ is the cumulative reward computed from a trajectory, $P(\tau | \pi_{\theta})$ is the probability of a trajectory given the current policy, and the angled brackets denote an average over many trajectories sampled under the $\pi_{\theta}$ policy.
This is achieved by encouraging the policy to increase the probability of selecting actions that returned good rewards and discouraging those that didn't.
In practice, training is done throughout discrete-time episodes where state, policy probabilities, and reward data are collected and used to update the policy via the gradient ascent algorithm
\begin{equation}
    \theta^{'} = \theta + \eta\cdot\nabla_{\theta}J(\pi_{\theta}),
\end{equation}
with learning rate $\eta$, and
\begin{equation}
    \nabla_{\theta}J = \left\langle \sum\limits_{t}^{T} \nabla_{\theta}\log \pi_{\theta}(a_{t}|s_{t})\cdot r_{t} \right\rangle_{\tau\sim\pi_{\theta}}
    \label{eqn:cost-gradient}
\end{equation}
with time step, $t$, and instantaneous reward, $r_{t}$.
Typically, to ensure convergence in the long-term rewards and to encourage near-term over long-term progress, an expected returns function, $G_{t}$, will be used in the form
\begin{equation}
    G_{t} = \sum\limits_{t^{'} = t}^{T}\gamma^{t^{'} - t}r_{t^{'}},
    \label{eqn:expected-returns}
\end{equation}
where $T$ is the final time step of the episode, and $\gamma$ is a discount factor set as a hyperparameter during training influencing the importance of late-time rewards.
While this approach can be sufficient in training agents, it can break down in stochastic environments and high-dimensional spaces~\citep{proconsvaluebasedrl}.
To address this,~\citet{sutton2018reinforcement} showed that introducing a trainable baseline or value function, $V^{\pi}$ to compute an advantage function
\begin{equation}
    A_{t}^{\pi} = G_{t} - V^{\pi}_{t},
\end{equation}
can result in the stabilization of the expected returns and improved convergence time in the training, resulting in the reformulation of Equation~\ref{eqn:cost-gradient} to
\begin{equation}
    \nabla_{\theta}J = \left\langle \sum\limits_{t}^{T} \nabla_{\theta}\log \pi_{\theta}(a_{t}|s_{t})\cdot A^{\pi}_{t} \right\rangle_{\tau\sim\pi_{\theta}}.
    \label{eqn:cost-gradient-improved}
\end{equation}
When this value function is implemented as a neural network, it is often referred to as a \texttt{critic}, thus actor-critic reinforcement learning.
The critic's role is to determine the value of the state in which the agent currently finds itself.
While several definitions exist, in \swarmrl{}, the value is defined as the theoretical returns possible if one starts in a state, $s_{t}$ and follows policy $\pi_{\theta}$. 
It improves this prediction by training on the true expected returns, Equation~\ref{eqn:expected-returns}, computed during episodes through a desired regression algorithm.
Therefore, in Equation~\ref{eqn:cost-gradient-improved}, the advantage function guides training by encouraging selected actions that are better than or equal to those expected from the value function and discouraging those considered below what is theoretically possible.
For this reason, AC RL training is often considered adversarial in nature, with the actor learning to beat the critic and the critic, in turn, learning the correct value of the encountered states.
The training procedure outlined here is referred to as Vanilla Policy Gradient due to it being the simplest form of actor-critic training.
More complex methods extend on the method discussed here, such as trust region policy optimisation~\citep{schulman17b} and proximal policy optimisation~\citep{schulman17a}, the latter of which is also implemented in \swarmrl{}.

\subsubsection{Multi-Agent Reinforcement Learning}
As \texttt{SwarmRL} specializes in controlling microscopic agents, there will be many occasions in which multiple agents must be simultaneously controlled and trained.
This is addressed by multi-agent reinforcement learning (MARL)~\citep{zhang2021multiagent}. 
In MARL, $N$ autonomous agents interact with the environment and each other. 
This extension brings several new aspects to the problem, such as cooperation and competition among agents and various information structures. 
In the following, we will address these aspects briefly. 
\paragraph{Cooperative, Competitive and Mixed Setting}
In a fully cooperative setting, all agents share a common reward, i.e., $r^1=r^2=...=r^N = r$ or a team-average reward where $r=\frac{1}{N}\sum_{i\in N}r_i$~\citep{zhang2018fully}. 
The latter is a generalized version of the former. 
Different reward functions for each agent allow more heterogeneity among agents~\citep{zhang2021multiagent}.
A fully competitive setting in MARL is typically modelled as a zero-sum Markov game with $\sum_{i\in N}r^i=0$~\citep{littman1994markov}. 
Most literature focuses on two players where one agent's reward is the other's loss, something leveraged in, for example, predator-prey style games.
A mixed setting imposes no restrictions on the goal and the relationship among agents~\citep{hu2003nash}. 
\paragraph{Information Structures}
Another critical challenge in MARL is the question \textit{who knows what} during training and execution. 
This involves both the decision-making process as well as the optimization process. 
Three different forms of information structure are generally accepted.
The first is the centralized setting with a so-called central controller aggregating joint observations, actions, and rewards and allowing for centralized learning. 
The execution, on the other hand, can either be centralized or decentralized.
A centralized executed policy produces a joint action for all agents, while decentralized produces 
individually executed policies. 
The latter is the popular learning scheme of centralized-learning-decentralized-execution~\citep{oliehoek2016concise,hansen2004dynamic}.
On the other side, there is the fully decentralized setting in which each agent acts based only on local observation. 
This setting results in the so-called independent learning scheme \citep{tan1993multi}.
A combination of both is a decentralized setting with networked agents~\citep{zhang2021multiagent}. 
This allows agents to share local information with their neighbours.
The choice of setting depends on the problem at hand. 
\section{Architecture}
\label{sec:architecture}
\begin{figure}[h]
    \centering
    \includegraphics[width=\linewidth]{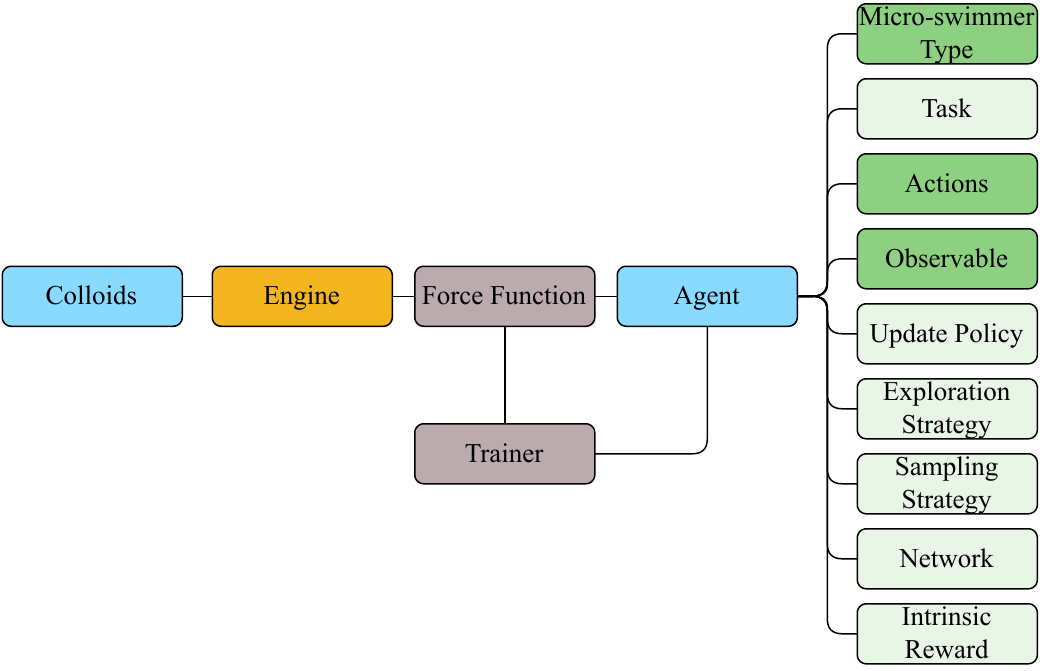}
    \caption{Overview of the \swarmrl{} software architecture. Light green boxes on the right-hand side of the figure represent modules with default settings that can be adjusted by users but do not need to be, dark green boxes indicate settings that need to be included in a system definition, blue boxes correspond to those directly handling colloid intelligence and properties, the engine is in orange, and grey represents classes that talk to the engine.}
    \label{fig:srl_flowchart}
\end{figure}

\swarmrl{} has been designed to lower the entry barrier for scientists to access state-of-the-art reinforcement learning algorithms efficiently.
It also offers a wide degree of customization and choice of algorithms that can be used during experiments to maximise the experienced user's flexibility.
To achieve both of these goals, \swarmrl{} is constructed using a modular, object oriented programming structure where almost all pipeline components are written as implementations of an abstract parent that defines the interface through which the components communicate.
The Python programming language has been used throughout as it is most common amongst scientific disciplines, however, whenever possible, libraries have been used to improve performance through vectorized maps (vmaps), multi-threading, GPU use, and Just In Time (JIT) compiling~\citep{bradbury18a}.
A coarse-grained flowchart of \swarmrl{} is presented in Figure~\ref{fig:srl_flowchart} where each major component of the architecture is shown with its corresponding connections.
The structure of \swarmrl{} is built around the interaction between the \texttt{Engine}, whether that be simulation or real experiment, and the \texttt{Agents}, the algorithms used to control the particles in the simulation.
Connecting the \texttt{Agents} to the \texttt{Engine} is the \texttt{Force Function}.
Force Functions are either constructed directly, or through a \texttt{Trainer} in the cases where trainable \texttt{Agents} are in use.
Finally, the \texttt{Agent} holds \texttt{Tasks}, \texttt{Actions}, and \texttt{Observables}, each of which provide information and functionality to the algorithm being used.
The rest of this section is dedicated to explaining each important component of \swarmrl{}, where they fit into the different capabilities of the software, and how they can be adapted or extended for individual implementations.
It should be noted that, to reduce space and improve readability, all comments, type hints, and methods not directly related to the discussion have been removed from code samples.
\subsection{Engine}
\label{sec:engine}
The \texttt{Engine} class of \swarmrl{} is one of its most important components.
It represents the environment of the \texttt{Agents} and propagates the state according to the given actions.
\swarmrl{} ships with an engine connected to the ESPResSo simulation software~\citep{weik19a} to solve Equations (\ref{eq:langevin_pos}) and (\ref{eq:langevin_direc}) numerically.
Through ESPResSo, we also offer the underdamped Langevin equation and a coupling to explicit hydrodynamics with lattice Boltzmann as the solver of the Navier-Stokes equations.
\swarmrl{} also ships with rudimentary examples for connecting it directly to an experiment. 
Direct experimental connections are often kept private out of respect to the institutions that host the experiments.
However, most implementations thus far use the TCP communications protocol to send actions and receive state information to hardware.
Future work on \swarmrl{} aims to unify this communication procedure using the ROS ecosystem~\citep{macenski22a}.
In cases where users must connect their own engines, be it simulation or experiment, they inherit from the \texttt{Engine} parent class. 
They must implement the two methods outline in Code Sample~\ref{lst:engine}.
\begin{lstlisting}[language=Python, label=lst:engine, caption=\swarmrl{} \texttt{Engine} parent class.]
class Engine:

    def integrate(
        self, n_slices, force_model,
    ):
        raise NotImplementedError

    def get_particle_data(self):
        raise NotImplementedError
        
\end{lstlisting}
The \texttt{integrate} method is used to step forward either in simulation or in an experiment.
It takes as arguments the force model, which holds the control algorithms for the particles in the experiment, see \cref{subsec:models}, and number of slices to be looped over, that is, the number of times the force model is called for new actions to be applied to the particles.
In a simulation, the time in-between successive time slices is filled with integrator steps, applying the chosen action at each time-step. 
In contrast, in experiment, this would be applying the actions for a fixed period between slices.
When instantiated, these parameters, time-step or time between actions, are given to the specific engine.
The second method that must be implemented is \texttt{get\_particle\_data}.
This method must collect information about the particles in the system and give it back to the user uniformly so that all parts of \swarmrl{} can work with them.
In its current state, \swarmrl{} defines colloids with properties outlined in Code Sample~\ref{lst:colloid}.
\begin{lstlisting}[language=Python, label=lst:colloid, caption=\swarmrl{} \texttt{Colloid} dataclass.]
class Colloid:
    pos
    director
    id
    velocity
    type
\end{lstlisting}
Here, all information relevant to the models is assigned to a colloid class and used to predict new actions and to compute rewards.
\subsection{Agents}
\label{subsec:agents}
\swarmrl{} was built to service a variety of simulations and RL studies.
A large component in enabling this, is the \texttt{Agent} class, which manages how agents in the simulations are controlled, what they see, and what they are tasked with doing.
The \texttt{Agent} collects the components introduced in the previous sections and represents the full set of sensing, control, and objectives of the colloids in question.
This information can then be used by \swarmrl{} to effectively deploy these agents in the simulation environment.
Code samples~\ref{lst:ac-agent} and~\ref{lst:class-agent} demonstrate the construction of two different agents which could both be added to a single simulation within \swarmrl{}.

\begin{lstlisting}[label=lst:ac-agent, caption=Actor-Critic Agent, language=Python]{Name}
ac_agent = srl.agents.ActorCriticAgent(
    agent_type=0,
    network=srl.networks.FlaxNetwork(...),
    task=srl.tasks.RotateRod(),
    observable=srl.observables.VisionCones(),
    actions=action_dict
)
\end{lstlisting}

\begin{lstlisting}[label=lst:class-agent, caption=Classical Agent, language=Python]{Name}
ac_agent = srl.agents.Lymburn(
    agent_type=1,
    homing_location=[500., 500., 500.]
)
\end{lstlisting}

In the above examples, the actor-critic reinforcement learning approach controls one group of particles. 
This means their policy can be trained and adapted to new environments.
The other example uses a so-called \texttt{ClassicalAgent}, in particular, one based on the swarming agents of~\citep{lymburn21a}.

\subsection{Classical Control}
\label{subsec:classical-control}
\swarmrl{}, allows users to control the actions of the particles in the system using both RL and classical algorithms.
Classical algorithms, in this work, can be implemented without training and typically follow a defined set of rules.
Such sets of rules have often been studied in the literature to reproduce behaviour in microswimmers~\citep{baeuerle20a, lavergne19a}.
Classical agents are implemented directly into the \texttt{Agent} child class on the software side. 
However, they must follow the same structure, namely, at each time slice, the information of all particles in the system is passed into the class and in turn, it must return an action for each particle.
In its current state, \swarmrl{} offers classical algorithms for the published models of~\citet{lymburn21a},~\citet{baeuerle20a}, and~\citet{lavergne19a}.
Future work aims at including Model Predictive Control (MPC) algorithms into the \swarmrl{} ecosystem.
\subsection{Machine Learning Control}
\label{subsec:machine-learning-control}
The machine learning side of agent control is driven by multi-agent reinforcement learning.
However, due to the complex nature of this approach, several design decision have been made to balance flexibility with limiting complexity.
The main workhorse of the machine learning driven control is the \texttt{Network} class, which requires two methods to be implemented, \texttt{compute\_action} which computes the action of the agent/s under study, and \texttt{\_\_call\_\_}, which returns action probabilities and, in the case of actor-critic approaches, also the predicted state values.
\subsubsection*{Architecture}
\swarmrl{} currently utilizes the Flax neural network library~\citep{heek23a} exclusively.
In order to provide an interface for complex architectures, \swarmrl{} requires only a single Flax module to be passed to the \texttt{FlaxNetwork} class.
This module must return both the action probabilities, and the predicted value function in case of actor-critic training.
This approach allows implementing very complex neural network architectures such as graph models, with the same API interface as simple dense actors and critics.
An example of two commonly used approaches is listed in Code Samples~\ref{lst:disjoint} and~\ref{lst:joint}.

\begin{lstlisting}[label=lst:disjoint, caption=Disjoint actor-critic, language=Python]{Name}
class DisjointActorCritic:

  def __call__(self, x):
    # Perform the actor forward pass.
    actor_values = Dense(12)(x)
    actor_values = relu(actor_values)
    actor_output = Dense(4)(actor_values)

    # Perform the critic forward pass.
    critic_values = Dense(12)(x)
    critic_values = relu(critic_values)
    critic_output = Dense(1)(critic_values)

        return actor_output, critic_output        
\end{lstlisting}

\begin{lstlisting}[label=lst:joint, caption=Combined actor-critic, language=Python]{Name}
class CombinedActorCritic:

  def __call__(self, x):
    # Perform the shared layer computation.
    shared_layer = Dense(12)(x)
    shared_layer = relu(shared_layer)

    # Compute the actor output.
    actor_output = Dense(4)(shared_layer)

    # Compute the critic output.
    critic_output = Dense(1)(shared_layer)

        return actor_output, critic_output      
\end{lstlisting}

These code blocks show a disjoint and combined actor critic approach bundled into the same Flax network structure.
In this way, many complex neural network architectures can be constructed under a single module including graph or transformer modules.

Apart from the architecture, several components are essential to training AC models: the policy update algorithm, the approach used in the expectation computation, and the strategy for sampling actions from the learned distribution.
\begin{table}[t]
\centering
\begin{tabular}{@{}c|ll@{}}
\toprule
\multirow{2}{*}{Policy Update}     & \multicolumn{2}{l}{Vanilla Gradient Update~\citep{sutton99a}}\\ \cmidrule(l){2-3} 
                                   & \multicolumn{2}{l}{Proximal Policy Optimization~\citep{schulman17a}}     \\ \midrule
\multirow{2}{*}{Returns}           & \multicolumn{2}{l}{Expected Returns}                                                           \\ \cmidrule(l){2-3} 
                                   & \multicolumn{2}{l}{Generalized Advantage Estimation~\citep{schulman18a}} \\ \midrule
\multirow{2}{*}{Sampling Strategy} & \multicolumn{2}{l}{Categorical Distribution}                                                   \\ \cmidrule(l){2-3} 
                                   & \multicolumn{2}{l}{Gumbel Trick~\citep{gumbel54a}}                       \\ \bottomrule
\end{tabular}
\caption{Overview of RL algorithm options available in \swarmrl{}}
\label{tab:alg-opt}
\end{table}

\subsubsection*{Actions}
Actions in \swarmrl{} are dictated by \texttt{dataclasses} which can be read by an \texttt{engine} and turned into forces, torques, or other values in a simulation or experiment.
At the time of this report, \swarmrl{} has been focused on micro-robotics with standard direction control, therefore, the class is built from a simple set of attributes outlined in Code Sample~\ref{lst:action}.
\begin{lstlisting}[language=Python, label=lst:action, caption=\swarmrl{} \texttt{Action} dataclass.]
class Action:
  force = 0.0
  torque= np.zeros((3,))
  new_direction = None
\end{lstlisting}
Force is applied along the agent's director axis, torque describes the agent's rotation, and new\_direction provides a sometimes simpler approach to a torque calculation.
From this template, we create an action by specifying the class's force, torque, and new direction attributes and add them to a dictionary to be passed to an agent.
For example, if the agent should be capable of translation forwards, translation backwards, clockwise, and counterclockwise rotation, the set of actions can be passed to the \texttt{Agent} as in Code Sample~\ref{lst:actionex}.
\begin{lstlisting}[language=Python, label=lst:actionex, caption=Example \texttt{Action} dictionary.]
agent_action_dictionary = {
  "Forwards": Action(force=10.0),
  "Backwards": Action(force=-10.0),
  "CCW": Action(torque=np.array(0., 0., 10.))
  "CW": Action(torque=np.array(0., 0., -10.))
}
\end{lstlisting}

\subsection{Force Function}
\label{subsec:models}
The interaction model in \swarmrl{} is responsible for consuming data from the simulation and sending back computed actions for each of the \texttt{agents}.
\swarmrl{} can control agents using machine learning and standard control theory approaches.
The parent class for the models requires only a single function to be called by the \texttt{Engine}: \texttt{calc\_action}, which takes a list of \texttt{agent} objects as input and returns a list of \texttt{actions}.
This is the only occasion in which the environment communicates with the control algorithms and the communication must adhere to this very 'narrow' interface.
This design choice enables novice \swarmrl{} users to switch out environments without adapting their RL setup and experienced users to easily implement new environments according to their needs.

Agent information is passed through as a dictionary when creating a model, demonstrated in Code Sample~\ref{lst:model}.
\begin{lstlisting}[language=Python, label=lst:model, caption=Example \texttt{ForceFunction} instantiation for a classical and actor-critic RL agent.]
interaction_model = ForceFunction(
    agents={
        "0": ActorCriticAgent,
        "1": ClassicalAgent
    }
)
\end{lstlisting}
In the case above, the experiment has at least two types of particles.
One is controlled using an actor-critic reinforcement learning approach while the other takes actions computed by a classical, non-trainable algorithm.
In this way, \swarmrl{} users have complete control over the agents in their simulations and experiments.

\subsection{Tasks}
\label{subsec:tasks}
As the goal of \swarmrl{} is to be used for the study of micro-scale robotic or biological systems to learn about their behaviour and control, it is often important to measure how well the system is doing.
In RL, this is often referred to as the reward which is maximised during training.
For classical algorithms, a task can also be used as a measure of how well the approach is doing or passed into the model as feedback.
In \swarmrl{}, system performance measurement is handled by the \texttt{Task} class.
The parent class of the task is built up of several methods that need to be implemented in the case of a custom task.
The reduced class architecture is displayed in Code Sample~\ref{lst:task}.
\begin{lstlisting}[language=Python, label=lst:task, caption=\swarmrl{} \texttt{Task} parent class.]
class Task:
  @property
  def kill_switch(self):
      pass
  def __call__(self, colloids):
      raise NotImplementedError("Implemented in child class.")
\end{lstlisting}
The above class contains two fundamental components, the kill switch and the call method.
As its name suggests, the kill switch is used to end the current run.
If this property is set to True, \swarmrl{} will stop the current running engine and either restart it, or just finish.
This is used, for example, in episodic training, where the training should stop and the engine should reset after an amount of time or when a criterion is reached.
However, the kill switch can also be used for more critical cases, such as in an experiment when it appears that the agents would perform damaging actions.
The second component, the call function, is used to compute a value describing the current state of this task, ideally, large positive values mean the task is being achieved, whereas negative values mean it is actively not being achieved. 
For example, when training RL agents to rotate an object, the call method would compute the rotation speed and return some positive value if it was increasing or large, and a negative value if it were decreasing.

At times, we want agents to perform more than just one task, in these cases, two options exist.
The first is to write a custom task that provides feedback on two different objects such as how fast something is rotating and how far it has been pushed to a desired location.
However, these tasks often quickly become convoluted and difficult to reuse.
Therefore, \swarmrl{} provides the \texttt{Multitasking} class which takes a list of single tasks as an argument and applies them to the engine.
This approach is more modular and allows for faster testing of strategies.

\subsection{Observables}
\label{subsec:observables}
A critical component of any RL algorithm and an important part of classical agent control approaches is their state description.
Rather than passing all state information (e.g., all positions and directions of the agents) to the decision making model, the observable usually selects from or condenses the information considerably.
This mimics the limited sensing capabilities that agents might have.
For example, agents might only be able to retrieve information about other particles within a finite sensing radius or might not be able to tell the direction of their neighbours.
Investigating the amount of information needed for an agent to still be able to perform its task is very relevant for the design of cost-efficient micro agents.
In \swarmrl{}, the description of the environment as sensed by the agents is computed by \texttt{Observables}.
All observables implemented in \swarmrl{} inherit from a simple parent class outlined in Code Sample~\ref{lst:observable}.
\begin{lstlisting}[language=Python, label=lst:observable, caption=\swarmrl{} \texttt{Observable} parent class.]
class Observable:
  def compute_observable(self, colloids):
    raise NotImplementedError("Implemented in child class.")
\end{lstlisting}
The class requires implementing one method, \texttt{compute\_observable}, which takes a list of particle objects as input and returns the state description vector for each of them.
This state vector is then passed on to the models by the force function.
As with the \texttt{Task} class, we often want to construct agents capable of sensing the world in multiple ways.
To achieve this, we also provide a \texttt{MultiSensing} class which takes a list of observables and computes a single output.

\subsection{Intrinsic Reward}
In RL agents can be rewarded intrinsically to either encourage the exploration behavior of an agent or to assist in building its knowledge about the environment~\citep{Pathak17a}. 
As the name suggests, such a method is intrinsic to the agent and is therefore implemented as a separate sub-module \texttt{IntrinsicReward} which the agent can be equipped with. 

\subsubsection*{Random Network Distillation}
Random network distillation (RND) was introduced in 2018 by~\citet{burda18a} as an efficient method for exploring the state space of reinforcement learning agents.
It has since been studied more broadly for its application and data selection for ML applications~\citep{finkbeiner23a} and its implications on understanding data requirements for neural networks~\citep{tovey23a}. \\
\swarmrl{} provides \texttt{RNDReward}, a module implementing RND, defining an intrinsic reward for agents, encouraging them to explore unseen regions of their state space. 
Going beyond the original version from~\citet{burda18a}, \swarmrl{} provides two implementations of RND, differing in their capability to memorize previous states.
For both versions of RND, a default setup is provided in \lstinline{swarmrl/intrinsic_reward/rnd_configs.py}. 
Utilizing such configuration the implementation of RND is shown in Code Sample~\ref{lst:intrinsic_reward_rnd}. 
\begin{lstlisting}[label=lst:intrinsic_reward_rnd, caption=Random Network Distillation as Intrinsic Reward, language=Python]{Name}
rnd_config = RNDConfig(...)

ac_agent = srl.agents.ActorCriticAgent(
    ...,
    intrinsic_reward=RNDReward(rnd_config)
) 
\end{lstlisting}
The implementation of the RND backend is based on the ZnNL python package \citep{nikolaou21a}, offering a flexible but simple interface for complex training algorithms for neural networks.

\subsection{Exploration Policy}
In addition to the sampling strategy which can select actions from its distributions that do not have the largest probability, \swarmrl{} also provides direct access to exploration policies to further guide the agents into new, unexplored regions of their environment.
\subsubsection*{Random Exploration}
The simplest form of exploration policy is the \texttt{RandomExploration} class, which an exploration probability can parameterize, $\zeta$ and decay rate, $\epsilon$.
The exploration rate dictates how likely it is for an agent to take an exploration action over a policy action, and the decay rate slowly decreases this probability as the training evolves by 
\begin{equation}
    \zeta^{'} = \exp^{-\epsilon\frac{t}{T}}\zeta,
\end{equation}
where $t$ is simulation time, and $T$ is the time for a single training episode.



\subsection{Trainer}
\label{subsec:trainer}
The trainer is module in \swarmrl{} responsible for handling the updates of the models and their interaction with the environment while developing a policy.
In the simplest case, the trainer deploys the models in the environment, collects the rewards over an episode, and performs a network update, as outlined in Code Sample~\ref{lst:cont-trainers}.
However, in many cases more is demanded during RL training and even how rewards are collected can change depending on the problem.
Therefore, in \swarmrl{}, we have implemented several variations of the \texttt{Trainer}.
\subsubsection*{Continuous vs Episodic}
A user must first choose whether to perform episodic or continuous training.
In episodic training, agents are deployed in the environment for a fixed number of steps or until some condition is met.
After this time, a reward is computed based on their final state or trajectory during the episode.
This reward is then used to update the networks before the environment is reset and the process is started again.
In this approach, the episodes should be long enough to allow the agents to achieve their tasks but short enough to maximize efficiency.
A similar but alternative approach is semi-episodic training, wherein the agents are updated continuously during the training but after some time, or again, once a condition is met, the environment is reset and the training continues.
These options are available through the \texttt{EpisodicTrainer} packaged into \swarmrl{}.
The \texttt{EpisodicTrainer}, displayed in Code Samples~\ref{lst:cont-trainers} and~\ref{lst:ep-trainers}, allows users to define how often the environment should be reset as a function of RL-episodes, or to pass a task which can force an environment reset.

\begin{lstlisting}[label=lst:cont-trainers, caption=Continuous Training, language=Python]{trainers}
rl_trainer = srl.trainers.ContinuousTrainer(
    [protocol_1, protocol_2],
    loss,
)
rl_trainer.perform_rl_training(
    system_runner=srl.Engine(...),
    n_episodes=5000,
    episode_length=50,
)
\end{lstlisting}

\begin{lstlisting}[label=lst:ep-trainers, caption=Episodic Training, language=Python]{trainers}
rl_trainer = srl.trainers.EpisodicTrainer(
                [protocol_1, protocol_2],
                loss,
            )
rl_trainer.perform_rl_training(
    get_engine=engine_fn,
    n_episodes=10,
    reset_frequency=30,  # > 1 is semi-episodic
    episode_length=100,
)     
\end{lstlisting}

Continuous training is a paradigm in which agents are deployed into an environment and are updated after a fixed amount of time without resetting the environment.
This online training is useful for situations where training must occur in the real world or where the time it will take to achieve a task is not well known.
\swarmrl{} offers the \texttt{ContinuousTrainer} class to perform this kind of training.
It should be noted that if a \texttt{Task} is provided to the continuous trainer capable of resetting the environment, the training will end.
This can also be helpful in cases where an end to training can be strictly identified.

\section{Performance}
\label{sec:performance}
Reinforcement learning typically does not require expensive neural networks, and classical control algorithms are limited in complexity by the number of particles they control.
However, speed becomes critical when considering the number of iterations they have to go through during training and deployment.
Therefore, \swarmrl{} has been built with a performance-first mentality, leveraging the JAX ecosystem~\citep{heek23a} heavily.
JAX is a Python library that, among other things, re-implements the NumPy library~\citep{harris20a} with multi-threading, GPU deployment, and XLA-enhanced JIT compile capabilities.
Most variables in \swarmrl{} are written using JAX arrays or their PyTree counterparts.
PyTrees extend the functionality JAX has for NumPy objects to data structures like classes.
Therefore, most of the class objects including the \texttt{Colloid} class can be placed onto a GPU or parallelized over.
Furthermore, large parts of code are JIT compiled to ensure optimal performance.

\section{Software Practices}
\label{sec:software-practices}
In order to ensure the longevity and continued usability of the \swarmrl{} software, we adhere to several software practices.
\swarmrl{} is thoroughly tested with unit, integration, and functional tests in place.
Overall, we have 87 \% coverage on the code with all classes having at least one unit test.
The code follows the NumPy doc-string format and is extensively documented which is hosted at \url{https://swarmrl.github.io/SwarmRL.ai/}.
Before new code is added to the project, all tests must pass and new tests should be included.
The project is hosted on GitHub where all these standards are enforced automatically.
Furthermore, we require at least one review from a code-owner, that is, a member of the core \swarmrl{} team, to accept new changes to the software.

\section{Visualization}
\label{sec:visualization}
\begin{figure}
    \centering
    \includegraphics[width=\linewidth]{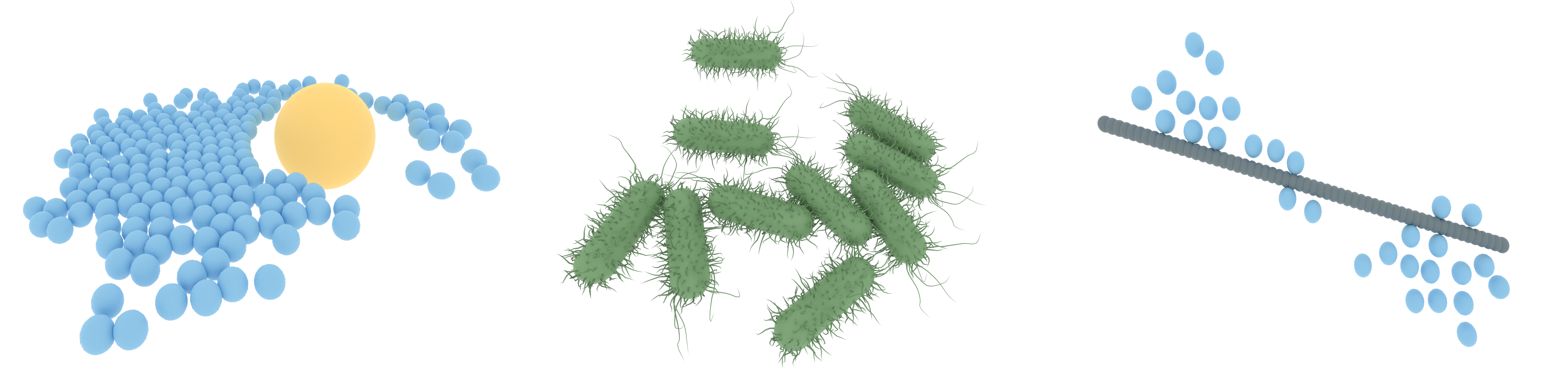}
    \caption{Example renderings using \swarmrl{} and the ZnVis visualization engine. (\textit{left}) A snapshot from a reservoir computing experiment using a reinforcement learning controlled swarm. (\textit{center}) Colloids are trained to perform chemotaxis in the same fashion as bacteria. Here an open-source bacteria 3d model file has been used inside of \swarmrl{} to convey realism. The mesh was created by Sketchfab artist andrewfrueh, whose explicit permission was requested and granted before being used in this publication. The model is licensed under creative commons 4.0. (\textit{right}) Demonstration of colloids rotating a rod, controlled using an RL algorithm.}
    \label{fig:znvis}
\end{figure}

Visualization is an essential part of the scientific process, particularly in simulations where one is searching for emergent strategy.
It is not always clear from plots alone if the agents have learned what they should do.
For this reason, \swarmrl{} interfaces closely with the ZnVis~\citep{tovey21a} visualization library.
ZnVis, by design, handles the output data from \swarmrl{} and is developed alongside the project.
The package can use custom mesh files for the agents, export rendered videos, and capture still-frame renderings using the Mitsuba engine~\citep{jakob22a}.
Figure~\ref{fig:znvis} illustrates the visualization capabilities of the package used on \swarmrl{} experiments.
\section{Conclusion}
\label{sec:conclusion}
We have introduced \swarmrl{}, a Python package aimed at accelerating research into control and understanding of microscopic active particles, whether biological or artificial.
\swarmrl{} allows researchers to implement state-of-the art control algorithms driven by reinforcement learning or classical policies, in both simulated and experiment settings.
As a Python package, \swarmrl{} has a familiar interface for many scientists, and is simple enough to use with limited background in software.
Beyond its functionality, we have built \swarmrl{} to be performant, capable of running on many cores, GPUs, and scalable to HPC clusters.
Our current focus is extending \swarmrl{} to handle more control algorithms, e.g., MPC, Q-learning, additional experiment implementations, and additional engines.
Overall, our goal is that \swarmrl{} allow scientists to spend more time creating new ideas, building new technology, and changing the field, rather than implementing and staying up to date with state of-of-the art machine learning methods.
\section{Acknowledgements}
C.H and S.T acknowledge financial support from the German Funding Agency (Deutsche Forschungsgemeinschaft DFG) under Germany’s Excellence Strategy EXC 2075-390740016.
C.H and S.T acknowledge financial support from the German Funding Agency (Deutsche Forschungsgemeinschaft DFG) under the Priority Program SPP 2363, “Utilization and Development of Machine Learning for Molecular Applications - Molecular Machine Learning” Project No. 497249646.
C.H and C.L acknowledge funding by the Deutsche Forschungsgemeinschaft (DFG, German Research Foundation) under Project Number 327154368-SFB 1313.
The authors acknowledge support by the state of Baden-Württemberg through bwHPC and the German Research Foundation (DFG) through grant INST 35/1597-1 FUGG.
The authors acknowledge support from the Deutsche Forschungsgemeinschaft (DFG, German Research Foundation) Compute Cluster grant no. 492175459.
S.T would like to acknowledge Annalena Daniels for her detailed review and comments on the manuscript.

\bibliography{bibliography}

\begin{thebibliography}{70}
\providecommand{\natexlab}[1]{#1}
\providecommand{\url}[1]{\texttt{#1}}
\expandafter\ifx\csname urlstyle\endcsname\relax
  \providecommand{\doi}[1]{doi: #1}\else
  \providecommand{\doi}{doi: \begingroup \urlstyle{rm}\Url}\fi

\bibitem[B{\"a}uerle et~al.(2020)B{\"a}uerle, L{\"o}ffler, and
  Bechinger]{baeuerle20a}
T.~B{\"a}uerle, R.~C. L{\"o}ffler, and C.~Bechinger.
\newblock Formation of stable and responsive collective states in suspensions
  of active colloids.
\newblock \emph{Nature Communications}, 11\penalty0 (1):\penalty0 2547, May
  2020.
\newblock ISSN 2041-1723.
\newblock \doi{10.1038/s41467-020-16161-4}.
\newblock URL \url{https://doi.org/10.1038/s41467-020-16161-4}.

\bibitem[Berg and Brown(1972)]{berg72a}
H.~C. Berg and D.~A. Brown.
\newblock Chemotaxis in escherichia coli analysed by three-dimensional
  tracking.
\newblock \emph{Nature}, 239\penalty0 (5374):\penalty0 500--504, 1972.

\bibitem[Blackiston et~al.(2021)Blackiston, Lederer, Kriegman, Garnier,
  Bongard, and Levin]{blackiston21a}
D.~Blackiston, E.~Lederer, S.~Kriegman, S.~Garnier, J.~Bongard, and M.~Levin.
\newblock A cellular platform for the development of synthetic living machines.
\newblock \emph{Science Robotics}, 6\penalty0 (52):\penalty0 eabf1571, 2021.
\newblock \doi{10.1126/scirobotics.abf1571}.
\newblock URL
  \url{https://www.science.org/doi/abs/10.1126/scirobotics.abf1571}.

\bibitem[Borra et~al.(2022)Borra, Biferale, Cencini, and Celani]{borra22a}
F.~Borra, L.~Biferale, M.~Cencini, and A.~Celani.
\newblock Reinforcement learning for pursuit and evasion of microswimmers at
  low reynolds number.
\newblock \emph{Phys. Rev. Fluids}, 7:\penalty0 023103, Feb 2022.
\newblock \doi{10.1103/PhysRevFluids.7.023103}.
\newblock URL \url{https://link.aps.org/doi/10.1103/PhysRevFluids.7.023103}.

\bibitem[Bradbury et~al.(2018)Bradbury, Frostig, Hawkins, Johnson, Leary,
  Maclaurin, Necula, Paszke, Vander{P}las, Wanderman-{M}ilne, and
  Zhang]{bradbury18a}
J.~Bradbury, R.~Frostig, P.~Hawkins, M.~J. Johnson, C.~Leary, D.~Maclaurin,
  G.~Necula, A.~Paszke, J.~Vander{P}las, S.~Wanderman-{M}ilne, and Q.~Zhang.
\newblock {JAX}: composable transformations of {P}ython+{N}um{P}y programs,
  2018.
\newblock URL \url{http://github.com/google/jax}.

\bibitem[Burda et~al.(2018)Burda, Edwards, Storkey, and Klimov]{burda18a}
Y.~Burda, H.~Edwards, A.~Storkey, and O.~Klimov.
\newblock Exploration by random network distillation, 2018.

\bibitem[Carlsen et~al.(2014)Carlsen, Edwards, Zhuang, Pacoret, and
  Sitti]{carlsen14a}
R.~W. Carlsen, M.~R. Edwards, J.~Zhuang, C.~Pacoret, and M.~Sitti.
\newblock Magnetic steering control of multi-cellular bio-hybrid microswimmers.
\newblock \emph{Lab on a Chip}, 14\penalty0 (19):\penalty0 3850--3859, 2014.

\bibitem[Dabbagh et~al.(2022)Dabbagh, Sarabi, Birtek, Seyfi, Sitti, and
  Tasoglu]{dabbagh22a}
S.~R. Dabbagh, M.~R. Sarabi, M.~T. Birtek, S.~Seyfi, M.~Sitti, and S.~Tasoglu.
\newblock 3d-printed microrobots from design to translation.
\newblock \emph{Nature Communications}, 13\penalty0 (1):\penalty0 5875, Oct
  2022.
\newblock ISSN 2041-1723.
\newblock \doi{10.1038/s41467-022-33409-3}.
\newblock URL \url{https://doi.org/10.1038/s41467-022-33409-3}.

\bibitem[Darnton et~al.(2007)Darnton, Turner, Rojevsky, and Berg]{darnton07a}
N.~C. Darnton, L.~Turner, S.~Rojevsky, and H.~C. Berg.
\newblock On torque and tumbling in swimming escherichia coli.
\newblock \emph{Journal of bacteriology}, 189\penalty0 (5):\penalty0
  1756--1764, 2007.

\bibitem[Degrave et~al.(2022)Degrave, Felici, Buchli, Neunert, Tracey,
  Carpanese, Ewalds, Hafner, Abdolmaleki, de~Las~Casas,
  et~al.]{deepmind_plasma}
J.~Degrave, F.~Felici, J.~Buchli, M.~Neunert, B.~Tracey, F.~Carpanese,
  T.~Ewalds, R.~Hafner, A.~Abdolmaleki, D.~de~Las~Casas, et~al.
\newblock Magnetic control of tokamak plasmas through deep reinforcement
  learning.
\newblock \emph{Nature}, 602\penalty0 (7897):\penalty0 414--419, 2022.

\bibitem[Dhatt-Gauthier et~al.(2023)Dhatt-Gauthier, Livitz, Wu, and
  Bishop]{dhatt23a}
K.~Dhatt-Gauthier, D.~Livitz, Y.~Wu, and K.~J.~M. Bishop.
\newblock Accelerating the design of self-guided microrobots in time-varying
  magnetic fields.
\newblock \emph{JACS Au}, 3\penalty0 (3):\penalty0 611--627, 2023.
\newblock \doi{10.1021/jacsau.2c00499}.
\newblock URL \url{https://doi.org/10.1021/jacsau.2c00499}.

\bibitem[Fawzi et~al.(2022)Fawzi, Balog, Huang, Hubert, Romera-Paredes,
  Barekatain, Novikov, R.~Ruiz, Schrittwieser, Swirszcz, Silver, Hassabis, and
  Kohli]{fawzi22a}
A.~Fawzi, M.~Balog, A.~Huang, T.~Hubert, B.~Romera-Paredes, M.~Barekatain,
  A.~Novikov, F.~J. R.~Ruiz, J.~Schrittwieser, G.~Swirszcz, D.~Silver,
  D.~Hassabis, and P.~Kohli.
\newblock Discovering faster matrix multiplication algorithms with
  reinforcement learning.
\newblock \emph{Nature}, 610\penalty0 (7930):\penalty0 47--53, 2022.
\newblock \doi{10.1038/s41586-022-05172-4}.
\newblock URL \url{https://doi.org/10.1038/s41586-022-05172-4}.

\bibitem[Felfoul et~al.(2016)Felfoul, Mohammadi, Taherkhani, de~Lanauze,
  Zhong~Xu, Loghin, Essa, Jancik, Houle, Lafleur, Gaboury, Tabrizian, Kaou,
  Atkin, Vuong, Batist, Beauchemin, Radzioch, and Martel]{felfoul16a}
O.~Felfoul, M.~Mohammadi, S.~Taherkhani, D.~de~Lanauze, Y.~Zhong~Xu, D.~Loghin,
  S.~Essa, S.~Jancik, D.~Houle, M.~Lafleur, L.~Gaboury, M.~Tabrizian, N.~Kaou,
  M.~Atkin, T.~Vuong, G.~Batist, N.~Beauchemin, D.~Radzioch, and S.~Martel.
\newblock Magneto-aerotactic bacteria deliver drug-containing nanoliposomes to
  tumour hypoxic regions.
\newblock \emph{Nature Nanotechnology}, 11\penalty0 (11):\penalty0 941--947,
  Nov 2016.
\newblock ISSN 1748-3395.
\newblock \doi{10.1038/nnano.2016.137}.
\newblock URL \url{https://doi.org/10.1038/nnano.2016.137}.

\bibitem[Finkbeiner et~al.(2023)Finkbeiner, Tovey, and Holm]{finkbeiner23a}
J.~Finkbeiner, S.~Tovey, and C.~Holm.
\newblock Generating minimal training sets for machine learned potentials,
  2023.

\bibitem[Gumbel(1954)]{gumbel54a}
E.~J. Gumbel.
\newblock Statistical theory of extreme values and some practical applications.
  lectures by emit j. gumbel. national bureau of standards, washington, 1954.
  51 pp. diagrams. 40 cents.
\newblock \emph{The Aeronautical Journal}, 58\penalty0 (527):\penalty0
  792–793, 1954.
\newblock \doi{10.1017/S0368393100099958}.

\bibitem[Hansen et~al.(2004)Hansen, Bernstein, and
  Zilberstein]{hansen2004dynamic}
E.~A. Hansen, D.~S. Bernstein, and S.~Zilberstein.
\newblock Dynamic programming for partially observable stochastic games.
\newblock In \emph{AAAI}, volume~4, pages 709--715, 2004.

\bibitem[Harris et~al.(2020)Harris, Millman, van~der Walt, Gommers, Virtanen,
  Cournapeau, Wieser, Taylor, Berg, Smith, Kern, Picus, Hoyer, van Kerkwijk,
  Brett, Haldane, del R{\'i}o, Wiebe, Peterson, G{\'e}rard-Marchant, Sheppard,
  Reddy, Weckesser, Abbasi, Gohlke, and Oliphant]{harris20a}
C.~R. Harris, K.~J. Millman, S.~J. van~der Walt, R.~Gommers, P.~Virtanen,
  D.~Cournapeau, E.~Wieser, J.~Taylor, S.~Berg, N.~J. Smith, R.~Kern, M.~Picus,
  S.~Hoyer, M.~H. van Kerkwijk, M.~Brett, A.~Haldane, J.~F. del R{\'i}o,
  M.~Wiebe, P.~Peterson, P.~G{\'e}rard-Marchant, K.~Sheppard, T.~Reddy,
  W.~Weckesser, H.~Abbasi, C.~Gohlke, and T.~E. Oliphant.
\newblock Array programming with numpy.
\newblock \emph{Nature}, 585\penalty0 (7825):\penalty0 357--362, Sep 2020.
\newblock ISSN 1476-4687.
\newblock \doi{10.1038/s41586-020-2649-2}.
\newblock URL \url{https://doi.org/10.1038/s41586-020-2649-2}.

\bibitem[Hartl et~al.(2021)Hartl, Hübl, Kahl, and Zöttl]{hartl21a}
B.~Hartl, M.~Hübl, G.~Kahl, and A.~Zöttl.
\newblock Microswimmers learning chemotaxis with genetic algorithms.
\newblock \emph{Proceedings of the National Academy of Sciences}, 118\penalty0
  (19):\penalty0 e2019683118, 2021.
\newblock \doi{10.1073/pnas.2019683118}.
\newblock URL \url{https://www.pnas.org/doi/abs/10.1073/pnas.2019683118}.

\bibitem[Heek et~al.(2023)Heek, Levskaya, Oliver, Ritter, Rondepierre, Steiner,
  and van {Z}ee]{heek23a}
J.~Heek, A.~Levskaya, A.~Oliver, M.~Ritter, B.~Rondepierre, A.~Steiner, and
  M.~van {Z}ee.
\newblock {F}lax: A neural network library and ecosystem for {JAX}, 2023.
\newblock URL \url{http://github.com/google/flax}.

\bibitem[Hosseinidoust et~al.(2016)Hosseinidoust, Mostaghaci, Yasa, Park,
  Singh, and Sitti]{hosseinidoust16a}
Z.~Hosseinidoust, B.~Mostaghaci, O.~Yasa, B.-W. Park, A.~V. Singh, and
  M.~Sitti.
\newblock Bioengineered and biohybrid bacteria-based systems for drug delivery.
\newblock \emph{Advanced Drug Delivery Reviews}, 106:\penalty0 27--44, 2016.
\newblock ISSN 0169-409X.
\newblock \doi{https://doi.org/10.1016/j.addr.2016.09.007}.
\newblock URL
  \url{https://www.sciencedirect.com/science/article/pii/S0169409X16302629}.
\newblock Biologically-inspired drug delivery systems.

\bibitem[Howse et~al.(2007)Howse, Jones, Ryan, Gough, Vafabakhsh, and
  Golestanian]{howse07a}
J.~R. Howse, R.~A. Jones, A.~J. Ryan, T.~Gough, R.~Vafabakhsh, and
  R.~Golestanian.
\newblock Self-motile colloidal particles: from directed propulsion to random
  walk.
\newblock \emph{Physical review letters}, 99\penalty0 (4):\penalty0 048102,
  2007.

\bibitem[Hsu et~al.(2016)Hsu, Wong-Foy, McCoy, Cowan, Marlow, Chavez,
  Kobayashi, Shockey, and Pelrine]{hsu16a}
A.~Hsu, A.~Wong-Foy, B.~McCoy, C.~Cowan, J.~Marlow, B.~Chavez, T.~Kobayashi,
  D.~Shockey, and R.~Pelrine.
\newblock Application of micro-robots for building carbon fiber trusses.
\newblock In \emph{2016 International Conference on Manipulation, Automation
  and Robotics at Small Scales (MARSS)}, pages 1--6, 2016.
\newblock \doi{10.1109/MARSS.2016.7561729}.

\bibitem[Hu and Wellman(2003)]{hu2003nash}
J.~Hu and M.~P. Wellman.
\newblock Nash q-learning for general-sum stochastic games.
\newblock \emph{Journal of machine learning research}, 4\penalty0
  (Nov):\penalty0 1039--1069, 2003.

\bibitem[Ibarz et~al.(2021)Ibarz, Tan, Finn, Kalakrishnan, Pastor, and
  Levine]{ibarz21a}
J.~Ibarz, J.~Tan, C.~Finn, M.~Kalakrishnan, P.~Pastor, and S.~Levine.
\newblock How to train your robot with deep reinforcement learning: lessons we
  have learned.
\newblock \emph{The International Journal of Robotics Research}, 40\penalty0
  (4–5):\penalty0 698–721, Jan. 2021.
\newblock ISSN 1741-3176.
\newblock \doi{10.1177/0278364920987859}.
\newblock URL \url{http://dx.doi.org/10.1177/0278364920987859}.

\bibitem[Jakob et~al.(2022)Jakob, Speierer, Roussel, and Vicini]{jakob22a}
W.~Jakob, S.~Speierer, N.~Roussel, and D.~Vicini.
\newblock {DR}.{JIT}.
\newblock \emph{{ACM} Transactions on Graphics}, 41\penalty0 (4):\penalty0
  1--19, jul 2022.
\newblock \doi{10.1145/3528223.3530099}.
\newblock URL \url{https://doi.org/10.1145%2F3528223.3530099}.

\bibitem[Jarrell et~al.(2021)Jarrell, Albers, and Machado]{jarrell21a}
K.~F. Jarrell, S.-V. Albers, and J.~N. d.~S. Machado.
\newblock A comprehensive history of motility and archaellation in archaea.
\newblock \emph{FEMS microbes}, 2:\penalty0 xtab002, 2021.

\bibitem[Jiang et~al.(2010)Jiang, Yoshinaga, and Sano]{jiang10a}
H.-R. Jiang, N.~Yoshinaga, and M.~Sano.
\newblock Active motion of a janus particle by self-thermophoresis in a
  defocused laser beam.
\newblock \emph{Physical review letters}, 105\penalty0 (26):\penalty0 268302,
  2010.

\bibitem[Khalil et~al.(2014)Khalil, Magdanz, Sanchez, Schmidt, and
  Misra]{khalil14a}
I.~S.~M. Khalil, V.~Magdanz, S.~Sanchez, O.~G. Schmidt, and S.~Misra.
\newblock Biocompatible, accurate, and fully autonomous: a sperm-driven
  micro-bio-robot.
\newblock \emph{Journal of Micro-Bio Robotics}, 9\penalty0 (3):\penalty0
  79--86, Aug 2014.
\newblock ISSN 2194-6426.
\newblock \doi{10.1007/s12213-014-0077-9}.
\newblock URL \url{https://doi.org/10.1007/s12213-014-0077-9}.

\bibitem[Kim et~al.(2016)Kim, Ali, Cheang, Jeong, Kim, and Kim]{kim16a}
H.~Kim, J.~Ali, U.~K. Cheang, J.~Jeong, J.~S. Kim, and M.~J. Kim.
\newblock Micro manipulation using magnetic microrobots.
\newblock \emph{Journal of Bionic Engineering}, 13\penalty0 (4):\penalty0
  515--524, 2016.
\newblock ISSN 1672-6529.
\newblock \doi{https://doi.org/10.1016/S1672-6529(16)60324-4}.
\newblock URL
  \url{https://www.sciencedirect.com/science/article/pii/S1672652916603244}.

\bibitem[Kiran et~al.(2021)Kiran, Sobh, Talpaert, Mannion, Sallab, Yogamani,
  and Pérez]{kiran21a}
B.~R. Kiran, I.~Sobh, V.~Talpaert, P.~Mannion, A.~A.~A. Sallab, S.~Yogamani,
  and P.~Pérez.
\newblock Deep reinforcement learning for autonomous driving: A survey, 2021.

\bibitem[Kriegman et~al.(2020)Kriegman, Blackiston, Levin, and
  Bongard]{kriegman20a}
S.~Kriegman, D.~Blackiston, M.~Levin, and J.~Bongard.
\newblock A scalable pipeline for designing reconfigurable organisms.
\newblock \emph{Proceedings of the National Academy of Sciences}, 117\penalty0
  (4):\penalty0 1853--1859, 2020.
\newblock \doi{10.1073/pnas.1910837117}.
\newblock URL \url{https://www.pnas.org/doi/abs/10.1073/pnas.1910837117}.

\bibitem[Lavergne et~al.(2019)Lavergne, Wendehenne, B{\"a}uerle, and
  Bechinger]{lavergne19a}
F.~A. Lavergne, H.~Wendehenne, T.~B{\"a}uerle, and C.~Bechinger.
\newblock Group formation and cohesion of active particles with visual
  perception-dependent motility.
\newblock \emph{Science}, 364\penalty0 (6435):\penalty0 70--74, Apr. 2019.

\bibitem[Li et~al.(2022)Li, Pal, Aghakhani, Pena-Francesch, and Sitti]{li22a}
M.~Li, A.~Pal, A.~Aghakhani, A.~Pena-Francesch, and M.~Sitti.
\newblock Soft actuators for real-world applications.
\newblock \emph{Nature Reviews Materials}, 7\penalty0 (3):\penalty0 235--249,
  Mar 2022.
\newblock ISSN 2058-8437.
\newblock \doi{10.1038/s41578-021-00389-7}.
\newblock URL \url{https://doi.org/10.1038/s41578-021-00389-7}.

\bibitem[Littman(1994)]{littman1994markov}
M.~L. Littman.
\newblock Markov games as a framework for multi-agent reinforcement learning.
\newblock In \emph{Machine learning proceedings 1994}, pages 157--163.
  Elsevier, 1994.

\bibitem[Lymburn et~al.(2021)Lymburn, Algar, Small, and Jüngling]{lymburn21a}
T.~Lymburn, S.~D. Algar, M.~Small, and T.~Jüngling.
\newblock {Reservoir computing with swarms}.
\newblock \emph{Chaos: An Interdisciplinary Journal of Nonlinear Science},
  31\penalty0 (3):\penalty0 033121, 03 2021.
\newblock ISSN 1054-1500.
\newblock \doi{10.1063/5.0039745}.
\newblock URL \url{https://doi.org/10.1063/5.0039745}.

\bibitem[Macenski et~al.(2022)Macenski, Foote, Gerkey, Lalancette, and
  Woodall]{macenski22a}
S.~Macenski, T.~Foote, B.~Gerkey, C.~Lalancette, and W.~Woodall.
\newblock Robot operating system 2: Design, architecture, and uses in the wild.
\newblock \emph{Science Robotics}, 7\penalty0 (66):\penalty0 eabm6074, 2022.
\newblock \doi{10.1126/scirobotics.abm6074}.
\newblock URL
  \url{https://www.science.org/doi/abs/10.1126/scirobotics.abm6074}.

\bibitem[Mandal et~al.(2018)Mandal, Patil, Kakoty, and Ghosh]{mandal18a}
P.~Mandal, G.~Patil, H.~Kakoty, and A.~Ghosh.
\newblock Magnetic active matter based on helical propulsion.
\newblock \emph{Accounts of chemical research}, 51\penalty0 (11):\penalty0
  2689--2698, 2018.

\bibitem[Medina-Sánchez et~al.(2016)Medina-Sánchez, Schwarz, Meyer,
  Hebenstreit, and Schmidt]{medina16a}
M.~Medina-Sánchez, L.~Schwarz, A.~K. Meyer, F.~Hebenstreit, and O.~G. Schmidt.
\newblock Cellular cargo delivery: Toward assisted fertilization by
  sperm-carrying micromotors.
\newblock \emph{Nano Letters}, 16\penalty0 (1):\penalty0 555--561, 2016.
\newblock \doi{10.1021/acs.nanolett.5b04221}.
\newblock URL \url{https://doi.org/10.1021/acs.nanolett.5b04221}.
\newblock PMID: 26699202.

\bibitem[Mnih et~al.(2015)Mnih, Kavukcuoglu, Silver, Rusu, Veness, Bellemare,
  Graves, Riedmiller, Fidjeland, Ostrovski, Petersen, Beattie, Sadik,
  Antonoglou, King, Kumaran, Wierstra, Legg, and Hassabis]{mnih15a}
V.~Mnih, K.~Kavukcuoglu, D.~Silver, A.~A. Rusu, J.~Veness, M.~G. Bellemare,
  A.~Graves, M.~Riedmiller, A.~K. Fidjeland, G.~Ostrovski, S.~Petersen,
  C.~Beattie, A.~Sadik, I.~Antonoglou, H.~King, D.~Kumaran, D.~Wierstra,
  S.~Legg, and D.~Hassabis.
\newblock Human-level control through deep reinforcement learning.
\newblock \emph{Nature}, 518\penalty0 (7540):\penalty0 529--533, 2015.
\newblock \doi{10.1038/nature14236}.
\newblock URL \url{https://doi.org/10.1038/nature14236}.

\bibitem[Muiños-Landin et~al.(2021)Muiños-Landin, Fischer, Holubec, and
  Cichos]{landin21a}
S.~Muiños-Landin, A.~Fischer, V.~Holubec, and F.~Cichos.
\newblock Reinforcement learning with artificial microswimmers.
\newblock \emph{Science Robotics}, 6\penalty0 (52):\penalty0 eabd9285, 2021.
\newblock \doi{10.1126/scirobotics.abd9285}.
\newblock URL
  \url{https://www.science.org/doi/abs/10.1126/scirobotics.abd9285}.

\bibitem[Murphy et~al.(2008)Murphy, Tadokoro, Nardi, Jacoff, Fiorini, Choset,
  and Erkmen]{murphy08a}
R.~R. Murphy, S.~Tadokoro, D.~Nardi, A.~Jacoff, P.~Fiorini, H.~Choset, and
  A.~M. Erkmen.
\newblock \emph{Search and Rescue Robotics}, pages 1151--1173.
\newblock Springer Berlin Heidelberg, Berlin, Heidelberg, 2008.
\newblock ISBN 978-3-540-30301-5.
\newblock \doi{10.1007/978-3-540-30301-5_51}.
\newblock URL \url{https://doi.org/10.1007/978-3-540-30301-5_51}.

\bibitem[Nachum et~al.(2017)Nachum, Norouzi, Xu, and
  Schuurmans]{proconsvaluebasedrl}
O.~Nachum, M.~Norouzi, K.~Xu, and D.~Schuurmans.
\newblock Bridging the gap between value and policy based reinforcement
  learning.
\newblock In I.~Guyon, U.~V. Luxburg, S.~Bengio, H.~Wallach, R.~Fergus,
  S.~Vishwanathan, and R.~Garnett, editors, \emph{Advances in Neural
  Information Processing Systems}, volume~30. Curran Associates, Inc., 2017.
\newblock URL
  \url{https://proceedings.neurips.cc/paper_files/paper/2017/file/facf9f743b083008a894eee7baa16469-Paper.pdf}.

\bibitem[Nelson and Pané(2023)]{nelson23a}
B.~J. Nelson and S.~Pané.
\newblock Delivering drugs with microrobots.
\newblock \emph{Science}, 382\penalty0 (6675):\penalty0 1120--1122, 2023.
\newblock \doi{10.1126/science.adh3073}.
\newblock URL \url{https://www.science.org/doi/abs/10.1126/science.adh3073}.

\bibitem[Nelson et~al.(2010)Nelson, Kaliakatsos, and Abbott]{nelson10a}
B.~J. Nelson, I.~K. Kaliakatsos, and J.~J. Abbott.
\newblock Microrobots for minimally invasive medicine.
\newblock \emph{Annual Review of Biomedical Engineering}, 12\penalty0
  (1):\penalty0 55--85, 2010.
\newblock \doi{10.1146/annurev-bioeng-010510-103409}.
\newblock URL \url{https://doi.org/10.1146/annurev-bioeng-010510-103409}.
\newblock PMID: 20415589.

\bibitem[Nikolaou and Tovey(2021)]{nikolaou21a}
K.~Nikolaou and S.~Tovey.
\newblock {ZnNL}, July 2021.
\newblock URL \url{https://github.com/zincware/ZnNL}.

\bibitem[Oliehoek et~al.(2016)Oliehoek, Amato, et~al.]{oliehoek2016concise}
F.~A. Oliehoek, C.~Amato, et~al.
\newblock \emph{A concise introduction to decentralized POMDPs}, volume~1.
\newblock Springer, 2016.

\bibitem[Pathak et~al.(2017)Pathak, Agrawal, Efros, and Darrell]{Pathak17a}
D.~Pathak, P.~Agrawal, A.~A. Efros, and T.~Darrell.
\newblock Curiosity-driven exploration by self-supervised prediction.
\newblock In \emph{2017 IEEE Conference on Computer Vision and Pattern
  Recognition Workshops (CVPRW)}, pages 488--489, 2017.
\newblock \doi{10.1109/CVPRW.2017.70}.

\bibitem[Qin et~al.(2023)Qin, Zou, Zhu, and Pak]{qin23a}
K.~Qin, Z.~Zou, L.~Zhu, and O.~S. Pak.
\newblock {Reinforcement learning of a multi-link swimmer at low Reynolds
  numbers}.
\newblock \emph{Physics of Fluids}, 35\penalty0 (3), 03 2023.
\newblock ISSN 1070-6631.
\newblock \doi{10.1063/5.0140662}.
\newblock URL \url{https://doi.org/10.1063/5.0140662}.
\newblock 032003.

\bibitem[Romanczuk et~al.(2012)Romanczuk, B{\"a}r, Ebeling, Lindner, and
  Schimansky-Geier]{romanczuk12a}
P.~Romanczuk, M.~B{\"a}r, W.~Ebeling, B.~Lindner, and L.~Schimansky-Geier.
\newblock Active brownian particles.
\newblock \emph{The European Physical Journal Special Topics}, 202\penalty0
  (1):\penalty0 1--162, Mar 2012.
\newblock ISSN 1951-6401.
\newblock \doi{10.1140/epjst/e2012-01529-y}.
\newblock URL \url{https://doi.org/10.1140/epjst/e2012-01529-y}.

\bibitem[Schmidt et~al.(2020)Schmidt, Medina-S{\'a}nchez, Edmondson, and
  Schmidt]{schmidt20a}
C.~K. Schmidt, M.~Medina-S{\'a}nchez, R.~J. Edmondson, and O.~G. Schmidt.
\newblock Engineering microrobots for targeted cancer therapies from a medical
  perspective.
\newblock \emph{Nature Communications}, 11\penalty0 (1):\penalty0 5618, Nov
  2020.
\newblock ISSN 2041-1723.
\newblock \doi{10.1038/s41467-020-19322-7}.
\newblock URL \url{https://doi.org/10.1038/s41467-020-19322-7}.

\bibitem[Schulman et~al.(2017{\natexlab{a}})Schulman, Levine, Moritz, Jordan,
  and Abbeel]{schulman17b}
J.~Schulman, S.~Levine, P.~Moritz, M.~I. Jordan, and P.~Abbeel.
\newblock Trust region policy optimization, 2017{\natexlab{a}}.

\bibitem[Schulman et~al.(2017{\natexlab{b}})Schulman, Wolski, Dhariwal,
  Radford, and Klimov]{schulman17a}
J.~Schulman, F.~Wolski, P.~Dhariwal, A.~Radford, and O.~Klimov.
\newblock Proximal policy optimization algorithms, 2017{\natexlab{b}}.

\bibitem[Schulman et~al.(2018)Schulman, Moritz, Levine, Jordan, and
  Abbeel]{schulman18a}
J.~Schulman, P.~Moritz, S.~Levine, M.~Jordan, and P.~Abbeel.
\newblock High-dimensional continuous control using generalized advantage
  estimation, 2018.

\bibitem[Shen et~al.(2023)Shen, Cai, Wang, Ge, and Yang]{shen23a}
H.~Shen, S.~Cai, Z.~Wang, Z.~Ge, and W.~Yang.
\newblock Magnetically driven microrobots: Recent progress and future
  development.
\newblock \emph{Materials \& Design}, 227:\penalty0 111735, 2023.
\newblock ISSN 0264-1275.
\newblock \doi{https://doi.org/10.1016/j.matdes.2023.111735}.
\newblock URL
  \url{https://www.sciencedirect.com/science/article/pii/S0264127523001508}.

\bibitem[Silflow and Lefebvre(2001)]{silflow01a}
C.~D. Silflow and P.~A. Lefebvre.
\newblock Assembly and motility of eukaryotic cilia and flagella. lessons from
  chlamydomonas reinhardtii.
\newblock \emph{Plant physiology}, 127\penalty0 (4):\penalty0 1500--1507, 2001.

\bibitem[Silver et~al.(2016)Silver, Huang, Maddison, Guez, Sifre, van~den
  Driessche, Schrittwieser, Antonoglou, Panneershelvam, Lanctot, Dieleman,
  Grewe, Nham, Kalchbrenner, Sutskever, Lillicrap, Leach, Kavukcuoglu, Graepel,
  and Hassabis]{silver16a}
D.~Silver, A.~Huang, C.~J. Maddison, A.~Guez, L.~Sifre, G.~van~den Driessche,
  J.~Schrittwieser, I.~Antonoglou, V.~Panneershelvam, M.~Lanctot, S.~Dieleman,
  D.~Grewe, J.~Nham, N.~Kalchbrenner, I.~Sutskever, T.~Lillicrap, M.~Leach,
  K.~Kavukcuoglu, T.~Graepel, and D.~Hassabis.
\newblock Mastering the game of go with deep neural networks and tree search.
\newblock \emph{Nature}, 529\penalty0 (7587):\penalty0 484--489, 2016.
\newblock \doi{10.1038/nature16961}.
\newblock URL \url{https://doi.org/10.1038/nature16961}.

\bibitem[Stefanec et~al.(2022)Stefanec, Hofstadler, Krajník, Turgut, Alemdar,
  Lennox, Şahin, Arvin, and Schmickl]{stefanec22a}
M.~Stefanec, D.~N. Hofstadler, T.~Krajník, A.~E. Turgut, H.~Alemdar,
  B.~Lennox, E.~Şahin, F.~Arvin, and T.~Schmickl.
\newblock A minimally invasive approach towards “ecosystem hacking” with
  honeybees.
\newblock \emph{Frontiers in Robotics and AI}, 9, 2022.
\newblock ISSN 2296-9144.
\newblock \doi{10.3389/frobt.2022.791921}.
\newblock URL
  \url{https://www.frontiersin.org/articles/10.3389/frobt.2022.791921}.

\bibitem[Su et~al.(2019)Su, {Hurd Price}, Jing, Tian, Liu, and Qian]{su19a}
H.~Su, C.-A. {Hurd Price}, L.~Jing, Q.~Tian, J.~Liu, and K.~Qian.
\newblock Janus particles: design, preparation, and biomedical applications.
\newblock \emph{Materials Today Bio}, 4:\penalty0 100033, 2019.
\newblock ISSN 2590-0064.
\newblock \doi{https://doi.org/10.1016/j.mtbio.2019.100033}.
\newblock URL
  \url{https://www.sciencedirect.com/science/article/pii/S2590006419300596}.

\bibitem[Sutton and Barto(2018)]{sutton2018reinforcement}
R.~S. Sutton and A.~G. Barto.
\newblock \emph{Reinforcement learning: An introduction}.
\newblock MIT press, 2018.

\bibitem[Sutton et~al.(1999)Sutton, McAllester, Singh, and Mansour]{sutton99a}
R.~S. Sutton, D.~McAllester, S.~Singh, and Y.~Mansour.
\newblock Policy gradient methods for reinforcement learning with function
  approximation.
\newblock In S.~Solla, T.~Leen, and K.~M\"{u}ller, editors, \emph{Advances in
  Neural Information Processing Systems}, volume~12. MIT Press, 1999.
\newblock URL
  \url{https://proceedings.neurips.cc/paper_files/paper/1999/file/464d828b85b0bed98e80ade0a5c43b0f-Paper.pdf}.

\bibitem[Tan(1993)]{tan1993multi}
M.~Tan.
\newblock Multi-agent reinforcement learning: Independent vs. cooperative
  agents.
\newblock In \emph{Proceedings of the tenth international conference on machine
  learning}, pages 330--337, 1993.

\bibitem[Tovey(2021)]{tovey21a}
S.~Tovey.
\newblock {ZnVis: A Visualisation Package for Particle Simulations}, 2021.

\bibitem[Tovey et~al.(2023)Tovey, Krippendorf, Nikolaou, and Holm]{tovey23a}
S.~Tovey, S.~Krippendorf, K.~Nikolaou, and C.~Holm.
\newblock Towards a phenomenological understanding of neural networks: data.
\newblock \emph{Machine Learning: Science and Technology}, 4\penalty0
  (3):\penalty0 035040, sep 2023.
\newblock \doi{10.1088/2632-2153/acf099}.
\newblock URL \url{https://dx.doi.org/10.1088/2632-2153/acf099}.

\bibitem[Wadhwa and Berg(2022)]{wadhwa22a}
N.~Wadhwa and H.~C. Berg.
\newblock Bacterial motility: machinery and mechanisms.
\newblock \emph{Nature reviews microbiology}, 20\penalty0 (3):\penalty0
  161--173, 2022.

\bibitem[Wang et~al.(2019)Wang, Chen, Alcântara, Sevim, Hoop, Terzopoulou,
  de~Marco, Hu, de~Mello, Falcaro, Furukawa, Nelson, Puigmartí-Luis, and
  Pané]{wang19a}
X.~Wang, X.-Z. Chen, C.~C.~J. Alcântara, S.~Sevim, M.~Hoop, A.~Terzopoulou,
  C.~de~Marco, C.~Hu, A.~J. de~Mello, P.~Falcaro, S.~Furukawa, B.~J. Nelson,
  J.~Puigmartí-Luis, and S.~Pané.
\newblock Mofbots: Metal–organic-framework-based biomedical microrobots.
\newblock \emph{Advanced Materials}, 31\penalty0 (27):\penalty0 1901592, 2019.
\newblock \doi{https://doi.org/10.1002/adma.201901592}.
\newblock URL
  \url{https://onlinelibrary.wiley.com/doi/abs/10.1002/adma.201901592}.

\bibitem[Weik et~al.(2019)Weik, Weeber, Szuttor, Breitsprecher, de~Graaf,
  Kuron, Landsgesell, Menke, Sean, and Holm]{weik19a}
F.~Weik, R.~Weeber, K.~Szuttor, K.~Breitsprecher, J.~de~Graaf, M.~Kuron,
  J.~Landsgesell, H.~Menke, D.~Sean, and C.~Holm.
\newblock Espresso 4.0--an extensible software package for simulating soft
  matter systems.
\newblock \emph{The European Physical Journal Special Topics}, 227:\penalty0
  1789--1816, 2019.

\bibitem[Zeng et~al.(2023)Zeng, Zhou, Rinaldi, Gneiting, and Nori]{zeng23a}
Y.~Zeng, Z.-Y. Zhou, E.~Rinaldi, C.~Gneiting, and F.~Nori.
\newblock Approximate autonomous quantum error correction with reinforcement
  learning.
\newblock \emph{Phys. Rev. Lett.}, 131:\penalty0 050601, Jul 2023.
\newblock \doi{10.1103/PhysRevLett.131.050601}.
\newblock URL \url{https://link.aps.org/doi/10.1103/PhysRevLett.131.050601}.

\bibitem[Zhang et~al.(2018)Zhang, Yang, Liu, Zhang, and Başar]{zhang2018fully}
K.~Zhang, Z.~Yang, H.~Liu, T.~Zhang, and T.~Başar.
\newblock Fully decentralized multi-agent reinforcement learning with networked
  agents, 2018.

\bibitem[Zhang et~al.(2021)Zhang, Yang, and Başar]{zhang2021multiagent}
K.~Zhang, Z.~Yang, and T.~Başar.
\newblock Multi-agent reinforcement learning: A selective overview of theories
  and algorithms, 2021.

\bibitem[Zhuang et~al.(2015)Zhuang, Wright~Carlsen, and Sitti]{zhuang15a}
J.~Zhuang, R.~Wright~Carlsen, and M.~Sitti.
\newblock ph-taxis of biohybrid microsystems.
\newblock \emph{Scientific Reports}, 5\penalty0 (1):\penalty0 11403, Jun 2015.
\newblock ISSN 2045-2322.
\newblock \doi{10.1038/srep11403}.
\newblock URL \url{https://doi.org/10.1038/srep11403}.

\end{thebibliography}

\end{document}